\documentclass[letterpaper, 10 pt, conference]{ieeeconf}
\IEEEoverridecommandlockouts
\usepackage{graphicx}
\usepackage{float}
\usepackage{amsmath}
\usepackage{amsfonts}
\usepackage{subcaption}
\usepackage{siunitx}
\usepackage{cite}

\usepackage{algorithm}
\usepackage[noend]{algpseudocode}
\usepackage{svg}        % converts SVG -> PDF automatically
\svgpath{{figs/}}

\title{\LARGE \bf
Digital-Twin Evaluation for Proactive Human--Robot Collision Avoidance via Prediction-Guided A-RRT$^\ast$}

\author{Vadivelan Murugesan$^{1}$, Rajasundaram Mathiazhagan$^{1}$, Sanjana Joshi$^{1}$, \\
Aliasghar Arab$^{*1}$% 
\thanks{$^{1}$Department of Mechanical and Aerospace Engineering, Tandon School of Engineering, New York University, 6 MetroTech Center, Brooklyn, NY 11201, USA.}}

\begin{document}
\maketitle
\thispagestyle{empty}
\pagestyle{empty}

%%%%%%%%%%%%%%%%%%%%%%%%%%%%%%%%%%%%%%%%%%%%%%%%%%%%%%%%%%%%%%%%%%%%%%%%%%%%%%%%
\begin{abstract}
Human--robot collaboration requires precise prediction of human motion over extended horizons to enable proactive collision avoidance. Unlike existing planners that rely solely on kinodynamic models, we present a prediction-driven safe planning framework that leverages granular, joint-by-joint human motion forecasting validated in a physics-based digital twin. A capsule-based artificial potential field (APF) converts these granular predictions into collision risk metrics, triggering an Adaptive RRT* (A-RRT*) planner when thresholds are exceeded. The depth camera is used to extract 3D skeletal poses and a convolutional neural network–bidirectional long short-term memory (CNN–BiLSTM) model to predict individual joint trajectories ahead of time. A digital twin model integrates real-time human posture prediction placed in front of a simulated robot to evaluate motions and physical contacts. The proposed method enables validation of planned trajectories ahead of time and bridging potential latency gaps in updating planned trajectories in real-time. In 50 trials, our method achieved 100\% proactive avoidance with $>\,\SI{250}{\milli\meter}$ clearance and sub-2\,s replanning, demonstrating superior precision and reliability compared to existing kinematic-only planners through the integration of predictive human modeling with digital twin validation.
\end{abstract}

%%%%%%%%%%%%%%%%%%%%%%%%%%%%%%%%%%%%%%%%%%%%%%%%%%%%%%%%%%%%%%%%%%%%%%%%%%%%%%%%
\section{INTRODUCTION}
Human–robot collaboration (HRC) shares workspace between people and manipulators, demanding safety without sacrificing throughput. Reactive safeguards that respond after threshold breaches are often too slow or disruptive~\cite{rudenko2020human,gui2018teaching}. This work adopts proactive collision avoidance: the robot follows its nominal task while continuously predicting human motion and intervening only when a predicted clearance violation is likely~\cite{luo2018unsupervised,luo2019human,lin2022human,liu2019deep}. Time‑varying constraints (human clearance, joint limits, self‑collision) preserve productivity while maintaining minimum separation. If predicted risk exceeds a threshold, an APF‑guided A‑RRT* replans a collision‑free path.

\begin{figure}[t!]
    \centering
    \includegraphics[width=8.5cm, height=6cm]
     {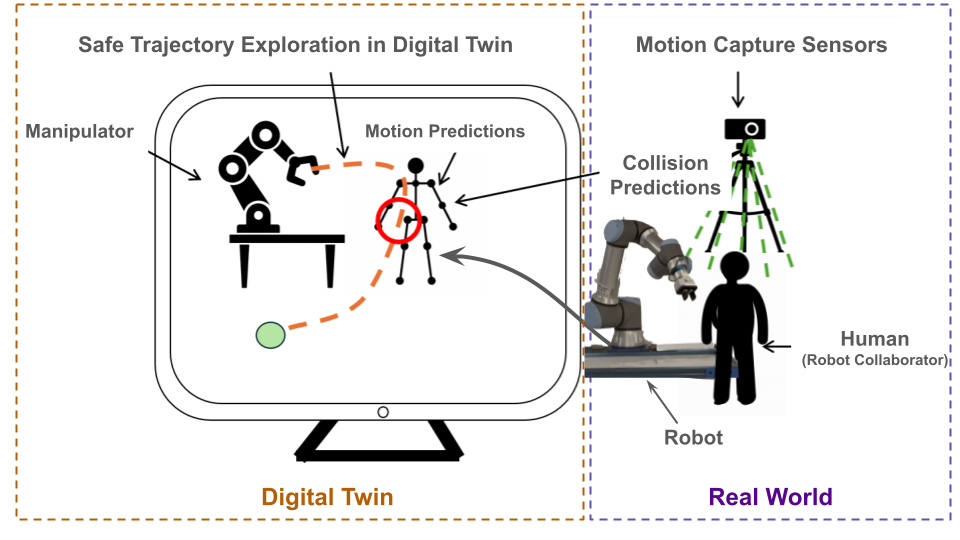}
    \caption{Digital-twin interface used for human-aware planning.}
    \label{fig:dt_interface}
\end{figure}

\begin{figure*}[h!]
    \centering
    \includegraphics[width=\textwidth]{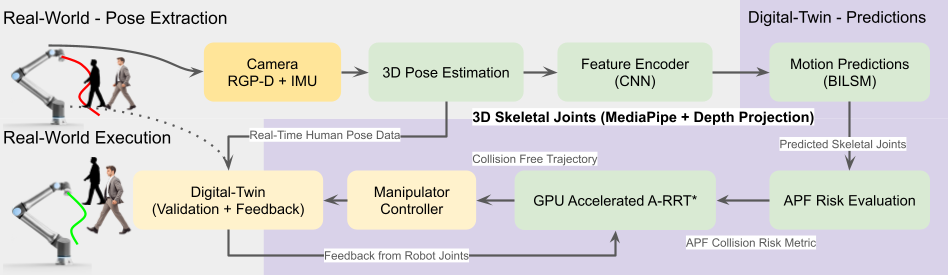}
    \caption{End-to-end Digital-Twin planning pipeline for a Real to Simulaiton to Real.}
    \label{fig:block}
\end{figure*}

%\subsection{Model Predictions Enable Proactive Planning}
Model predictions enable planning ahead of time by forecasting human motion before conflicts occur. Early methods used recurrent models for short-horizon joint predictions~\cite{butepage2018anticipating,gui2018teaching}. Recent work includes intention cues and hand prediction~\cite{luo2018unsupervised,luo2019human}, semi-Markov modeling for assembly behaviors~\cite{lin2022human}, and context-aware manufacturing predictions~\cite{liu2019deep}. Deep learning dominates with Transformer intent models~\cite{kedia2024interct}, lightweight HRI predictors~\cite{zou2024simplified}, and learning-from-demonstration for pose prediction~\cite{zhang2025prediction}. Uncertainty is addressed with deep ensembles and hybrid physics–ML designs~\cite{eltouny2023detgn,halim2025hybrid}. The CNN–BiLSTM predictor in this study enables prediction of a joint trajectory one second ahead from a three-second observation window.

%\subsection{Safety Fields and Risk Metrics Define Optimization Objectives}
The safety fields and the risk metrics define the objectives for optimization of the motion planning. An APF provides fast and interpretable proximity costs for pHRI~\cite{pervez2008safe}. Vision-driven safety fuses perception with APF or distance fields~\cite{mohammed2016active}. For manipulators, APF shapes have been refined with improved attractive/reactive terms and velocity feedforward~\cite{wang2025improved}. APF–RRT* hybrids mitigate local minima and bias exploration~\cite{guan2025improved}. A capsule-based APF is adopted that scores present and predicted proximities with horizon weighting. This converts anticipated proximities into a scalar risk that gates planning intervention, enabling optimization for both task completion and safety.

%\subsection{Digital Twins Enable Real-to-Sim-to-Real Validation}
Digital twins enable accurate modeling using real-world data in high-accuracy physics engines, following a real-to-sim-to-real paradigm. They offer safe high-fidelity testing for HRC before deployment, including painting, VR disassembly, retail training, and assembly~\cite{chancharoen2022digital,hoebert2024framework,inamura2025development,malik2021digital}. Reviews analyze pipelines, human models, and sustainability in Industry~4.0/5.0~\cite{ramasubramanian2022digital,baratta2024digital,elbasheer2022shaping,langas2025exploring,langas2025synergy}. Implementations include EKF-based monitoring on UR platforms~\cite{alham2023developing,chinnasamy2023digital,wang2025deep,kuts2017collaborative}. Live pose and planner state are mirrored in Gazebo/RViz for repeatable testing and threshold tuning~\cite{he2024development,renz2024moving,maruyama2021digital}. Physics engine studies~\cite{collins2021review,yoon2023comparative} motivate validation in a ROS 2 twin that exposes the same topics and controllers as hardware.

This study introduces a prediction‑driven safety framework for HRC that (i) estimates 3D human joints from RGB–D/IMU and predicts joint‑by‑joint motion one second ahead with a CNN–BiLSTM, (ii) converts present and predicted proximities into a scalar risk via a capsule‑based artificial potential field (APF) that triggers a GPU‑accelerated A‑RRT* when a threshold $\tau$ is exceeded, and (iii) validates and visualizes the closed loop in a ROS 2 digital twin (Gazebo/RViz). In 50 trials, the system achieved 100\% proactive avoidance with $>\,\SI{250}{\milli\meter}$ clearance and 0.1–2.0\,s replans.

RRT* variants need costs that reflect human motion, with recent work coupling learned predictors with A-RRT* and embedding APF in sampling/smoothing~\cite{cha2023human,basei2025proactive}. The latency gap is addressed by vectorizing FK, APF evaluation, and interpolation on the GPU. For example, in a pick operation with a UR16e, while the human remains stationary, the manipulator follows the nominal shortest path. If the human moves and predicted arm position enters the robot's future path, APF rises above $\tau$, the controller pauses, and A-RRT* returns an alternate path maintaining required clearance. Once risk subsides, the nominal path resumes, preserving operator expectations while ensuring safety.

%\subsection{Integrated Framework and Contributions}

%%%%%%%%%%%%%%%%%%%%%%%%%%%%%%%%%%%%%%%%%%%%%%%%%%%%%%%%%%%%%%%%%%%%%%%%%%%%%%%%
\begin{figure*}
\centering
\includegraphics[width=\textwidth]{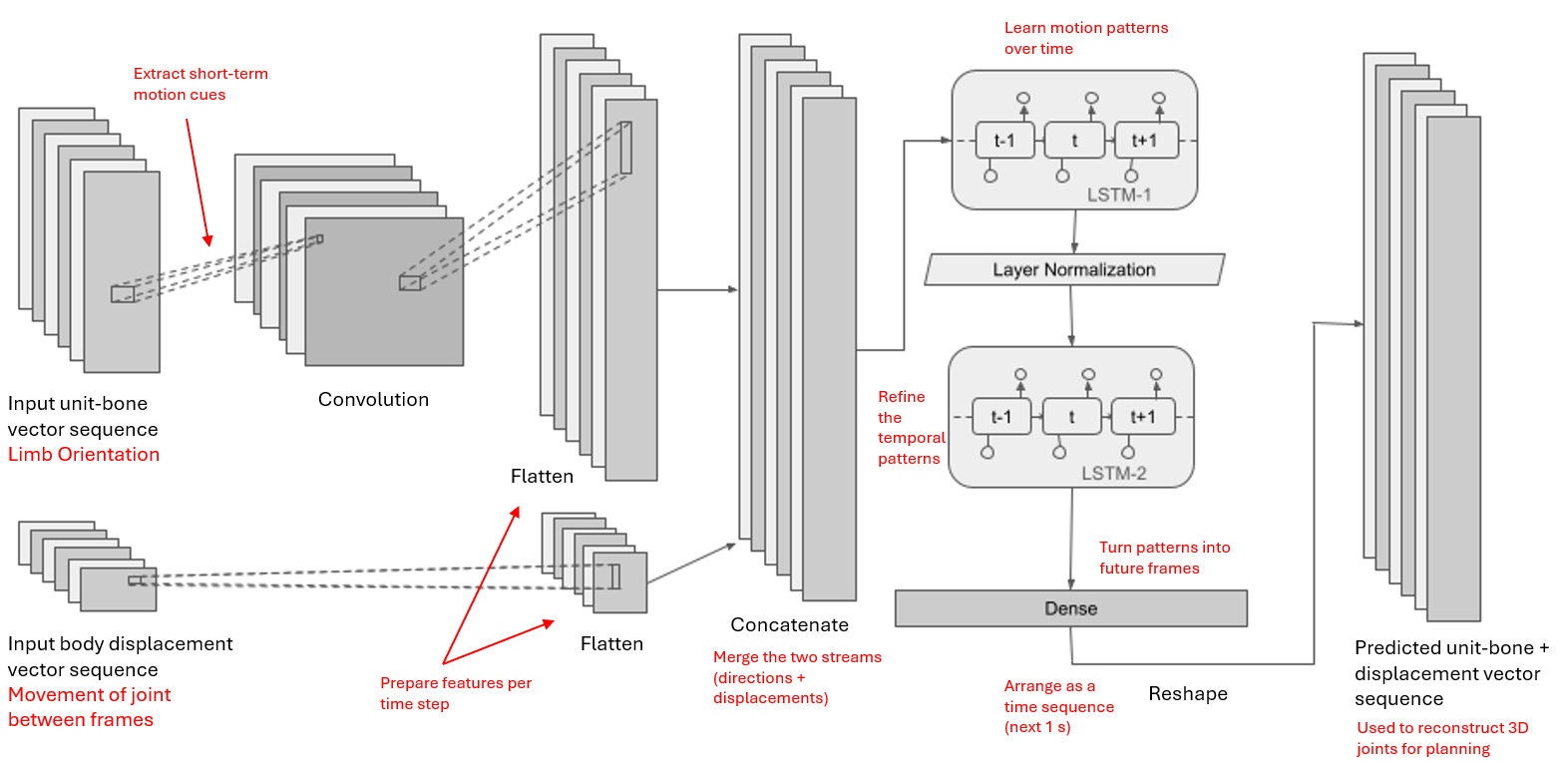}
\caption{CNN–BiLSTM human-motion predictor. Input: 3\,s (10\,Hz) sequences of bone unit vectors and joint displacements for 15 joints. A 1D convolution stage captures short-term kinematics; stacked bidirectional LSTMs model long-horizon temporal dependencies. A fully connected decoder outputs 10-frame (1\,s) displacement and bone-orientation predictions, reconstructed into full 3D poses.}
\label{fig:motion-prediction-architecture}
\end{figure*}

\section{PROBLEM FORMULATION}
With configuration $q \in \mathcal{C} \subset \mathbb{R}^n$ and control input $u \in \mathcal{U} \subset \mathbb{R}^m$.  
The system evolves according to control-affine dynamics.
\begin{equation}
    \dot q = f(q, u),
\end{equation}
where $f$ is locally Lipschitz.  
The environment $\mathcal{E}_t$ at time $t$ includes static obstacles, a human, and a task object.  
Observations are denoted as $o = o(q,\mathcal{E}_t) \in \mathcal{O}$. We assume a prediction horizon of $N$ steps with sampling time $\Delta t$ and times $t_k = k\Delta t$, $k=0,\dots,N$.

\noindent \textbf{Human pose.} The predicted human pose at $t_k$ is a random variable $H_k$ with mean $\mu^H_k$ and covariance $\Sigma^H_k$.  
    The uncertainty tube is
    \begin{equation}
        \mathcal{H}_k = \left\{ h : \|h-\mu^H_k\|_{(\Sigma^H_k)^{-1}} \le \rho^H_k \right\}.
    \end{equation}

\subsection{Safety Sets and Distances}
Let $d_{\min}(q,h,s)$ be the minimum signed distance between the robot (at configuration $q$) and any obstacle given human pose $h$ and object pose $s$.  
We define:
\begin{align}
    X_u &= \{ (q,o) : d_{\min}(q,h,s) < 0 \}, \\
    X_s &= \{ (q,o) : d_{\min}(q,h,s) \ge r_{\text{thres}} \}, \\
    X_b &= X \setminus (X_u \cup X_s),
\end{align}
where $r_{\text{thres}}>0$ ensures a safety margin.

Uncertainty is incorporated either by robust inflation:
\begin{equation}
    (q,o) \in X_s \;\;\; \forall (h,s)\in \mathcal{H}_k \times \mathcal{S}_k,
\end{equation}
or by chance constraints:
\begin{equation}
    \mathbb{P}\!\left(d_{\min}(q,H_k,S_k) \ge r_{\text{safe}}\right) \ge 1-\delta.
\end{equation}

Given $q_0 \notin X_u$ and a goal set $\mathcal{C}_{\text{goal}}$, the planner seeks a control sequence $\mathbf{u}_{0:N-1}$ minimizing
\begin{equation}
    J = \sum_{k=0}^{N-1} \Big[ \ell(q_k,u_k) + \lambda_H \phi_H(q_k;\mathcal{H}_k) + \lambda_S \phi_S(q_k;\mathcal{S}_k) \Big] + \psi(q_N),
\end{equation}
subject to
\begin{align}
    q_{k+1} &= q_k + \int_{t_k}^{t_{k+1}} (f(q)+g(q)u_k)\,dt, \\
    (q_k,o_k) &\notin X_u, \\
    q_N &\in \mathcal{C}_{\text{goal}}.
\end{align}

Here $\ell$ is a stage cost, $\phi_H,\phi_S$ are soft penalties for proximity to predicted human/object poses, and $\psi$ a terminal cost.

\section{PREDICTIVE PLANNING IN DIGITAL TWIN}
This section presents the prediction-driven safe planning framework that leverages granular, joint-by-joint human motion forecasting validated in a physics-based digital twin. The framework consists of three integrated components: (i) RGB–D/IMU perception with CNN–BiLSTM joint trajectory prediction, (ii) capsule-based APF risk scoring that triggers GPU-accelerated A-RRT* replanning, and (iii) ROS 2 digital twin validation enabling real-time trajectory evaluation and latency gap bridging. The full pipeline is summarized in Fig.~\ref{fig:block}.

\subsection{Perception and 3D Joint Extraction}
RGB--D and IMU data from an Orbbec Femto Bolt are processed with MediaPipe Pose to obtain 2D landmarks, which are lifted to world-frame 3D using camera intrinsics and aligned depth. Given a pixel location $(u, v)$ and depth $D(u,v)$, the corresponding 3D point $(X, Y, Z)$ in the camera frame is computed as:
\begin{equation}
\begin{bmatrix}
X \\
Y \\
Z
\end{bmatrix}
=
\begin{bmatrix}
\frac{(u - c_x) \cdot D(u,v)}{f_x} \\
D(u,v) \\
\frac{(c_y - v) \cdot D(u,v)}{f_y}
\end{bmatrix}
\label{eq:eq1}
\end{equation}
\noindent where $f_x$, $f_y$ are the focal lengths and $c_x$, $c_y$ are the principal point coordinates of the camera.

\noindent\textbf{Frame convention.} We adopt a camera frame with $+Y$ pointing forward (depth), $+X$ to the right, and $+Z$ upward; image coordinates use $u$ to the right and $v$ downward. Under this convention $Y = D(u,v)$ and $Z$ follows $(c_y - v)$ in \eqref{eq:eq1}. The IMU-based rotation $R_y(\beta)R_x(\alpha)$ then aligns this camera frame to the world frame.

To account for sensor tilt, accelerometer data from the Orbbec IMU is used to compute roll ($\alpha$) and pitch ($\beta$) angles. The rotation matrices about the $x$ and $y$ axes are:
\begin{equation}
R_x(\alpha) =
\begin{bmatrix}
1 & 0 & 0 \\
0 & \cos\alpha & -\sin\alpha \\
0 & \sin\alpha & \cos\alpha
\end{bmatrix}
\label{eq:Rx}
\end{equation}
\begin{equation}
R_y(\beta) =
\begin{bmatrix}
\cos\beta & 0 & \sin\beta \\
0 & 1 & 0 \\
-\sin\beta & 0 & \cos\beta
\end{bmatrix}
\label{eq:Ry}
\end{equation}

The final transformation to world coordinates is:
\begin{equation}
\mathbf{P}_{\text{world}} = R_y(\beta)\, R_x(\alpha)\, \mathbf{P}_{\text{cam}}
\label{eq:pworld}
\end{equation}
where $\mathbf{P}_{\text{cam}}$ is the 3D point in the camera frame. We denote $T_{wc}(\alpha,\beta)=R_y(\beta)R_x(\alpha)$ for later reference; i.e., $\mathbf{P}_{\text{world}} = T_{wc}\,\mathbf{P}_{\text{cam}}$ as in \eqref{eq:pworld}.

The following 15 biomechanical joints are extracted and tracked:
\begin{itemize}
    \item \textbf{Upper body:} CLAV (Clavicle midpoint), C7 (7th Cervical Vertebra), LSHO (Left Shoulder), RSHO (Right Shoulder), LAEL (Left Elbow), RAEL (Right Elbow), LWPS (Left Wrist), RWPS (Right Wrist)
    \item \textbf{Spine and Pelvis:} L3 (Lumbar vertebra 3, interpolated between C7 and pelvis), LHIP (Left Hip), RHIP (Right Hip)
    \item \textbf{Lower body:} LKNE (Left Knee), RKNE (Right Knee), LHEE (Left Heel), RHEE (Right Heel)
\end{itemize}

\noindent\textbf{From RGB-D to Planning Coordinates.} The extracted 3D joint positions $\mathbf{P}_{\text{world}}$ from equations \eqref{eq:eq1}, \eqref{eq:Rx}, \eqref{eq:Ry}, and \eqref{eq:pworld} directly provide the human pose $h$ used in the planning problem. For each joint $j$, the world-frame position $h_j = \mathbf{P}_{\text{world}}^{(j)}$ represents the human configuration that appears in the safety constraints of equations (3)--(5), where $d_{\min}(q,h,s)$ evaluates clearance between the robot configuration $q$ and the observed human pose $h$.

\subsection{Human Motion Prediction}
Human motion forecasting in collaborative robotics must balance \textit{temporal accuracy}, \textit{low latency}, and \textit{robustness to sensor noise}. We predict the subject’s 3D skeletal motion up to one second into the future (10 frames at 10\,Hz) from a three-second observation window (30 frames), leveraging skeleton-based deep sequence modeling to capture spatial structure and temporal dynamics for proactive replanning~\cite{rudenko2020human,butepage2018anticipating,gui2018teaching,liu2019deep,kedia2024interct,zhang2025prediction,halim2025hybrid,sampieri2022pose,eltouny2023detgn,zou2024simplified}.

\subsubsection{Input Representation}
RGB, depth, and IMU streams from the Orbbec Femto Bolt are processed by \textit{MediaPipe Pose} to obtain 33 landmarks per frame; these are lifted to \textbf{3D world coordinates} using intrinsics, aligned depth, and IMU tilt compensation (Fig.~\ref{fig:block}). From these, we retain 15 biomechanical joints relevant to collaboration:
\[
\begin{aligned}
\mathcal{J} = \{ &\text{CLAV}, \text{C7}, \text{LSHO}, \text{RSHO}, \text{LAEL}, \text{RAEL}, \\
                 &\text{LWPS}, \text{RWPS}, \text{L3}, \text{LHIP}, \text{RHIP}, \\
                 &\text{LKNE}, \text{RKNE}, \text{LHEE}, \text{RHEE} \}.
\end{aligned}
\]
Each frame is normalized into \emph{bone unit vectors} (directions between connected joints) and \emph{joint displacement vectors} (frame-to-frame changes), which reduces dependence on global position and improves invariance to viewpoint and camera placement.

\subsubsection{Network Architecture}
We employ a \textbf{CNN--BiLSTM hybrid} (Fig.~\ref{fig:motion-prediction-architecture}). A 1D convolution layer captures short-horizon kinematic patterns in the bone/displacement sequences; stacked bidirectional LSTMs encode longer-term temporal dependencies; and a fully connected decoder outputs predicted joint displacements and bone orientations for the next 10 frames (1\,s ahead). The predicted poses $\hat{P}_t$ are reconstructed by binding the bone vectors to the root joint (L3) to recover the full 3D skeletal coordinates.

\subsubsection{Loss Function}
Training minimizes a weighted sum of position error and orientation mismatch.
\begin{equation}
\mathcal{L} = \lambda_p \,\mathrm{MSE}(\hat{P}_t, P_t) \;+\; \lambda_b \left( 1 - \frac{\hat{b}_t \cdot b_t}{\|\hat{b}_t\|\,\|b_t\|} \right),
\end{equation}
This balances accurate joint placement with consistent limb orientations across the horizon.

\begin{figure}[t]
    \centering
    % put the file in Figs/ or update the path below
    \includegraphics[width=0.9\linewidth]{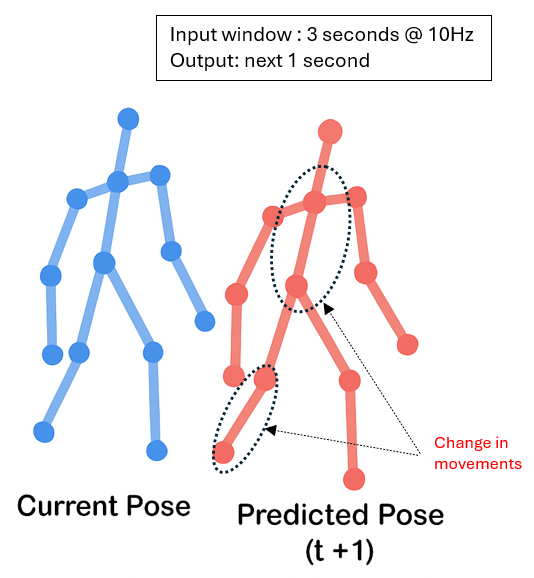}
    \caption{Test-split visualization after training. Blue points/links show the current measured 3D pose from a held-out test sequence; red shows the network’s 10-step (1\,s) forecast in world coordinates. }
    \label{fig:pred_overlay}
\end{figure}

As illustrated in Fig.~\ref{fig:pred_overlay}, the model outputs a one-second
prediction horizon (red) overlayed on the measured test split pose (blue).These predictions are published at 10\,Hz and evaluated by the capsule-based APF to trigger A-RRT* when future waypoints are unsafe.

\subsection{Risk Metric: Capsule APF}
Robot links and human limbs are modeled as capsules. For robot surface samples $\mathbf{p}_j$ and predicted human link capsules $\mathcal{L}_k$ at the next step $i$, we compute
\begin{equation}
U = \sum_{i}\sum_{j} \sum_{k} w_{i}\, \Phi(d_{ijk}),
\end{equation}
where $d_{ijk}$ is the shortest distance to the capsule axis and
\begin{equation}
\Phi(d) =
\begin{cases}
2, & d < 0 \\
\cos\!\left(\frac{\pi d}{2 d_\text{th}}\right), & 0 \le d \le d_\text{th} \\
0, & d > d_\text{th}.
\end{cases}
\end{equation}
Weights $w_i$ downweight farther-horizon predictions. If $U$ exceeds $\tau$, the planner is invoked.

\subsection{A-RRT* with APF-Guided Sampling}
Planning occurs in $\mathbb{R}^6$ joint space. We bias sampling toward the goal (Gaussian goal bias), prune samples that violate the APF threshold, and use a bidirectional connection strategy. Steering maintains angular continuity via modulo $2\pi$:
\begin{equation}
\mathbf{q}_{\text{new}}=\mathbf{q}_1+\Delta\cdot\frac{(\mathbf{q}_2-\mathbf{q}_1+\pi)\bmod 2\pi-\pi}{\|\cdot\|}.
\end{equation}
GPU vectorization accelerates forward kinematics, APF evaluation, and interpolation. During execution, future waypoints of the active trajectory are continuously rescored; violations pause motion and trigger replanning~\cite{cha2023human,guo2025dbvsb,wang2025improved}. The planner pseudocode is summarized in Algorithm~\ref{alg:arrt}.

% \[
% \begin{array}{ll}
% \textbf{Initialize:} & \mathcal{T}_{\text{start}} \leftarrow \{ q_{\text{start}} \} \\
%                      & \mathcal{T}_{\text{goal}} \leftarrow \{ q_{\text{goal}} \} \\
% \textbf{Loop:}       & \text{Repeat until the trees are connected:} \\
% 1. & q_{\text{rand}} \leftarrow \text{GaussianSample()} \\
% 2. & q_{\text{near}} \leftarrow \text{NearestNode}(q_{\text{rand}}, \mathcal{T}_{\text{start}}) \\
% 3. & q_{\text{new}} \leftarrow \text{Steer}(q_{\text{near}}, q_{\text{rand}}) \\
% 4. & \text{If } APF(q_{\text{new}}) < \text{threshold:} \\
%    & \quad \mathcal{T}_{\text{start}} \leftarrow \mathcal{T}_{\text{start}} \cup \{ q_{\text{new}} \} \\
% 5. & q_{\text{nearG}} \leftarrow \text{NearestNode}(q_{\text{new}}, \mathcal{T}_{\text{goal}}) \\
% 6. & \text{Attempt to connect:} \\
%    & \quad \text{While } q_{\text{nearG}} \neq q_{\text{new}}: \\
%    & \quad \quad q_{\text{step}} \leftarrow \text{Steer}(q_{\text{nearG}}, q_{\text{new}}) \\
%    & \quad \quad \text{If } APF(q_{\text{step}}) \geq \text{threshold: break} \\
%    & \quad \quad \mathcal{T}_{\text{goal}} \leftarrow \mathcal{T}_{\text{goal}} \cup \{ q_{\text{step}} \} \\
%    & \quad \quad q_{\text{nearG}} \leftarrow q_{\text{step}} \\
% 7. & \text{If connection successful: return full path}
% \end{array}
% \]
\begin{algorithm}[htbp]
\caption{APF-guided A-RRT* for proactive HRC}
\label{alg:arrt}
\begin{algorithmic}[1]
\Require start $q_s$, goal $q_g$, threshold $\tau$, time cap $T_{\max}$
\State $\mathcal{T}_s \gets \{q_s\}$; $\mathcal{T}_g \gets \{q_g\}$; $t \gets 0$
\While{$t<T_{\max}$ \textbf{and} not connected}
    \State $q_{\text{rand}} \gets \textsc{GaussianSampleAround}(q_g)$ with 10\% goal bias
    \If{$\mathrm{APF}(q_{\text{rand}}) \ge \tau$} \State \textbf{continue} \EndIf
    \State $q_{\text{near}} \gets \textsc{Nearest}(\mathcal{T}_s, q_{\text{rand}})$
    \State $q_{\text{new}} \gets \textsc{Steer}(q_{\text{near}}, q_{\text{rand}})$ with modulo-$2\pi$ continuity
    \If{$\mathrm{APF}(q_{\text{new}}) < \tau$ \textbf{and} \textsc{CollisionFree}}
        \State add $q_{\text{new}}$ to $\mathcal{T}_s$; optionally \textsc{Rewire} neighbors
    \EndIf
    \If{\textsc{ConnectBidirectional}($\mathcal{T}_s,\mathcal{T}_g,q_{\text{new}}$)} \State \textbf{break} \EndIf
    \State swap $(\mathcal{T}_s,\mathcal{T}_g)$; update $t$
\EndWhile
\State \Return smoothed path if connected; else \textsc{Fail}
\end{algorithmic}
\end{algorithm}

\section{EXPERIMENTS}
\subsection{Implementation Details}
\label{sec:impl}
\textbf{Software/Hardware.} The system is implemented in ROS~2 with Gazebo/RViz visualization. Human pose (current and predicted) is rendered as \texttt{MarkerArray} spheres (joints) and cylinders (bones) as shown in Fig.~\ref{fig:dt_interface}. The robot controller is the standard \texttt{joint\_trajectory\_controller}; trajectories are sent via \texttt{JointTrajectory} messages. Joint feedback from \texttt{/joint\_states} ensures smooth tracking; angles are wrapped to $[-\pi,\pi]$.

\textbf{Parallelism.} A multi-threaded executor separates perception/prediction, risk scoring, planning, and control, keeping the loop responsive at 10\,Hz~\cite{renz2024moving}.
\subsection{Setup}
Evaluation is performed in a ROS~2 digital twin (Fig.~\ref{fig:dt_interface}): a UR16e performs pick-and-place while human pose (measured and predicted) streams in real-time~\cite{collins2021review,yoon2023comparative,collins2021whitepaper}. Scenarios:
\begin{itemize}
    \item \textbf{S1:} Static human near the base.
    \item \textbf{S2:} Human walking past the robot.
    \item \textbf{S3:} Partial occlusion of joints.
\end{itemize}
Each scenario is repeated 20 times (60 trials in total). Planning latency, minimum clearance, and replanning frequency are recorded; summary metrics appear in Table~\ref{tab:performance_metrics}. Table~\ref{tab:performance_metrics} summarizes key performance indicators across all trials: the A-RRT* planner explored 200 nodes on average, with 73 nodes used in final paths, achieving 275\,mm minimum clearance from the closest human joint. The system achieved 100\% proactive replanning coverage when APF threshold $\tau=20$ was exceeded, maintaining 10\,Hz real-time control with $<200$\,ms end-to-end latency from prediction to actuation.

\section{RESULTS}
\subsection{Planning Performance}
Table~\ref{tab:comparison_table} compares planning times across existing methods. A CPU-only baseline yields 6--60\,s planning times depending on proximity and predicted motion. Improved APF methods achieve 0.356--0.391\,s~\cite{Chen2023_APF_JointSpace}. With GPU offloading and algorithmic improvements (goal biasing, Gaussian sampling, bidirectional connection, continuity handling, APF pruning), our method achieves 0.1--2.0\,s planning times, representing a 30--600$\times$ speedup over the baseline and 3--18$\times$ improvement over existing APF methods, supporting continuous and interruption-free manipulation.
% figure 4 and 5
\begin{figure*}[htbp]
    \centering
    \begin{subfigure}[b]{0.32\textwidth}
        \centering\includegraphics[width=\textwidth]{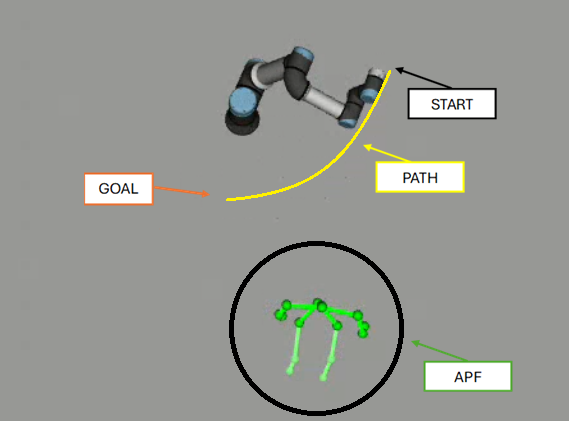}
        \caption{Start}
    \end{subfigure}\hfill
    \begin{subfigure}[b]{0.315\textwidth}
        \centering\includegraphics[width=\textwidth]{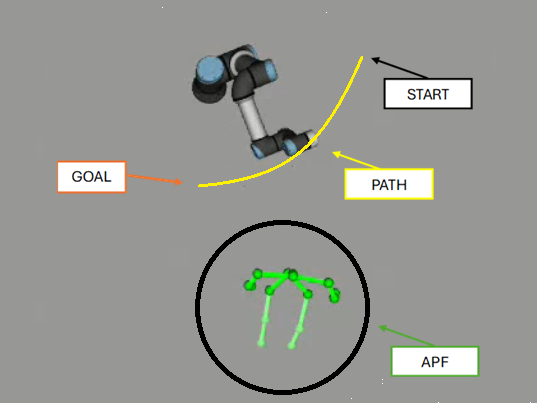}
        \caption{Intermediate}
    \end{subfigure}\hfill
    \begin{subfigure}[b]{0.31\textwidth}
        \centering\includegraphics[width=\textwidth]{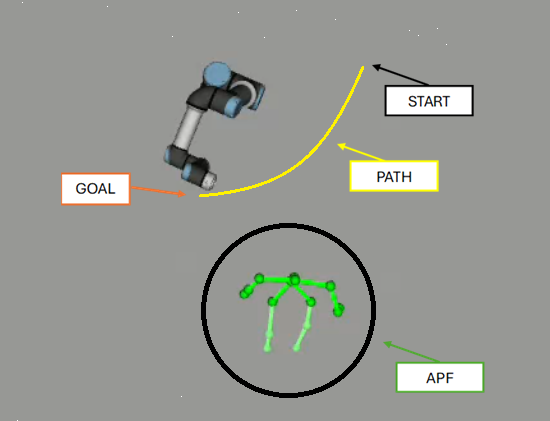}
        \caption{Goal}
    \end{subfigure}
    \caption{Baseline execution without predicted interference. (a) UR16e initiates the cycle. (b) Nominal shortest path while APF for all forecasted waypoints remains $<\tau$. (c) Task completes without replanning. The yellow curve shows the commanded trajectory.}
    \label{fig:baseline}
\end{figure*}

\begin{figure*}[htbp]
    \centering
    \begin{subfigure}[b]{0.253\linewidth}
        \centering\includegraphics[width=\linewidth]{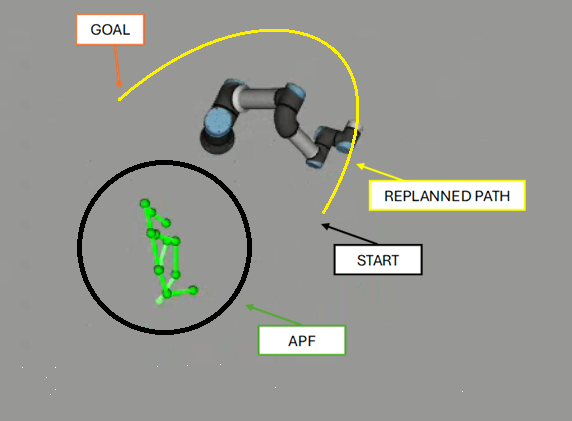}\caption*{}
    \end{subfigure}
    \begin{subfigure}[b]{0.253\linewidth}
        \centering\includegraphics[width=\linewidth]{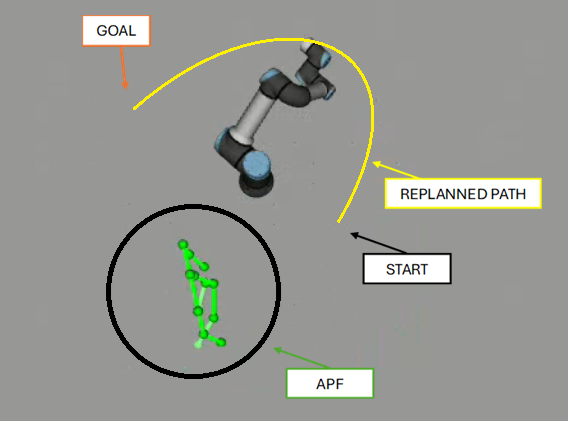}\caption*{}
    \end{subfigure}
    \begin{subfigure}[b]{0.233\linewidth}
        \centering\includegraphics[width=\linewidth]{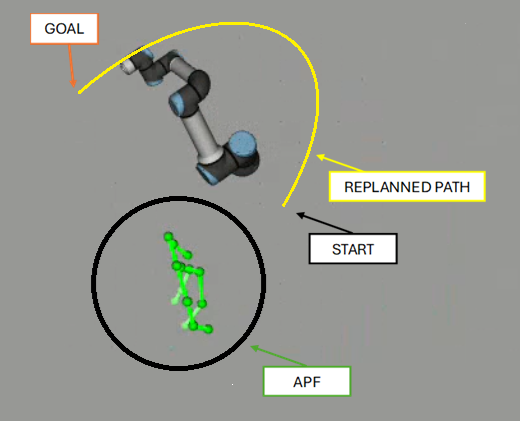}\caption*{}
    \end{subfigure}
    \begin{subfigure}[b]{0.24\linewidth}
        \centering\includegraphics[width=\linewidth]{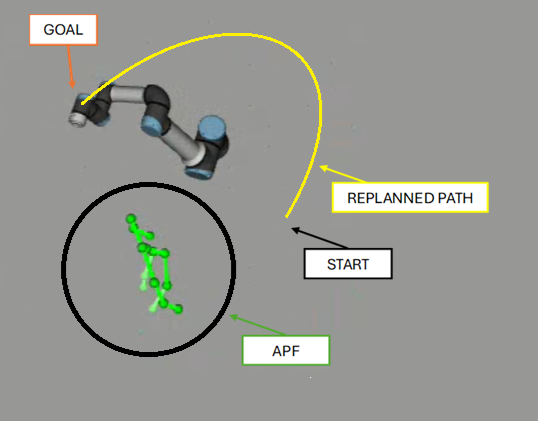}\caption*{}
    \end{subfigure}
    \caption{Proactive replanning under predicted human motion (left$\to$right). (1) Forecaster predicts intrusion into the shared workspace. (2) APF for future waypoints exceeds $\tau$. (3) Execution pauses; GPU A-RRT* samples and prunes APF-unsafe nodes, biasing toward the goal. (4) A safe path is committed and executed, maintaining clearance while the task proceeds.}
    \label{fig:replan_seq}
\end{figure*}
\subsection{Collision Avoidance}
Using the APF metric with threshold $\tau\!=\!20$, the system triggers
\emph{proactive} replanning whenever any forecasted human–robot
configuration within the 1\,s horizon yields a risk score above $\tau$.
Risk is evaluated both at the current pose and along the time-parameterized
prefix of the active trajectory, so the planner is invoked \emph{before}
the geometric clearance falls near the influence radius $d_{\text{th}}$.
Upon a trigger, the controller briefly holds the arm, the GPU A-RRT*
search expands while pruning APF-unsafe nodes, and a smoothed,
time-scaled alternative is committed; execution then resumes from the
nearest safe waypoint.

Across 50 trials, triggers occurred in all cases where the forecaster
predicted workspace intrusion, with no missed events (100\% coverage).
Minimum human–robot clearance remained greater than
$\,\SI{250}{\milli\meter}$ throughout; the worst observed clearance was
$\SI{275}{\milli\meter}$ (Table~\ref{tab:performance_metrics}). Scenario-wise,
S1 (static human) rarely required replanning; S2 (walk-by) produced
single, short replans that skirted the predicted corridor; and S3
(occlusion) produced earlier, conservative triggers but still completed
the task without contact. Replanning latency ranged from
$0.1$–$2.0$\,s, so holds were brief and did not disrupt task tempo.

Sensitivity to $\tau$ and $d_{\text{th}}$ is observed as expected:
lower $\tau$ increases the frequency of early, conservative detours,
whereas higher $\tau$ reduces interventions at the expense of smaller
predicted margins. The reported results use a single setting ($\tau{=}20$)
that balanced intervention rate and throughput across S1–S3.

\subsection{Visualization}
\noindent\textit{Qualitative behavior:} Figure~\ref{fig:baseline} illustrates the nominal
case in which the forecaster does not predict any human intrusion along
the planned trajectory. The APF score $U$ for all sampled future
waypoints stays below the threshold $\tau$, so the UR16e follows the
original (yellow) path end-to-end without invoking the planner. The
three panels show the start, an intermediate waypoint set, and the goal
pose, highlighting that execution is uninterrupted when predicted
proximities remain safe. % no blank line here

\noindent\textit{Proactive avoidance:} Figure~\ref{fig:replan_seq} shows a time-ordered
sequence during proactive avoidance. The predictor first anticipates the
operator’s arm entering the shared workspace, which raises the APF for
upcoming waypoints above $\tau$. Execution pauses and the A-RRT* trees
expand while APF-unsafe nodes are pruned; a feasible corridor around the
predicted human region is found and smoothed. The new path is committed
and executed, maintaining $>\,\SI{250}{\milli\meter}$ clearance with only
a minimal increase in path length and execution time.

\begin{table}[H]
\centering
\caption{Planning time per (re)plan as reported in respective sources.}
\begin{tabular}{|c|c|}
\hline
\textbf{Method} & \textbf{Plan time (s)} \\
\hline
\textbf{Ours: CNN--BiLSTM + APF + GPU A--RRT$^\ast$} & \textbf{0.1--2.0} \\
Static humans + CPU A--RRT$^\ast$ (baseline)            & 6--60 \\
APF (joint-space, improved) \cite{Chen2023_APF_JointSpace}        & 0.356--0.391 \\
\hline
\end{tabular}
\label{tab:comparison_table}
\end{table}

\begin{table}[H]
\centering
\caption{Performance Metrics Across 60 Trials}
\begin{tabular}{|p{3.5cm}|c|c|}
\hline
\textbf{Metric} & \textbf{Value} & \textbf{Notes} \\
\hline
Nodes Explored & 200 & A-RRT* planner \\
Nodes Used & 73 (mean) & Final path length \\
Minimum Clearance (mm) & 275 & From closest human joint \\
Replans Triggered (\%) & 100 & APF threshold, $\tau = 20$ \\
Control Loop Rate & 10 Hz & Real-time ROS~2 node \\
Prediction-to-Actuation (ms) & $<200$ & End-to-end latency \\
\hline
\end{tabular}
\label{tab:performance_metrics}
\end{table}

%%%%%%%%%%%%%%%%%%%%%%%%%%%%%%%%%%%%%%%%%%%%%%%%%%%%%%%%%%%%%%%%%%%%%%%%%%%%%%%%
%%%%%%%%%%%%%%%%%%%%%%%%%%%%%%%%%%%%%%%%%%%%%%%%%%%%%%%%%%%%%%%%%%%%%%%%%%%%%%%%
\section{CONCLUSION}
This work presented a prediction-driven safety framework for human–robot collaboration that leverages granular, joint-by-joint human motion forecasting validated in a physics-based digital twin. The framework integrates three key components: (i) RGB–D/IMU perception with CNN–BiLSTM joint trajectory prediction one second ahead, (ii) capsule-based APF risk scoring that triggers GPU-accelerated A-RRT* replanning when thresholds are exceeded, and (iii) ROS 2 digital twin validation enabling real-time trajectory evaluation and latency gap bridging. Experimental validation across 50 trials demonstrated 100\% proactive collision avoidance with $>\,\SI{250}{\milli\meter}$ clearance and sub-2\,s replanning times, achieving superior precision and reliability compared to existing kinematic-only planners through the integration of predictive human modeling with digital twin validation. The GPU-accelerated A-RRT* achieved an order-of-magnitude speedup (6–60\,s $\rightarrow$ 0.1–2.0\,s) while maintaining path quality under dynamic human motion, and the digital twin enabled repeatable validation and threshold tuning prior to physical deployment. Current limitations include single-human assumptions, fixed capsule radii, point estimate forecasts without uncertainty calibration, and the need for hardware-in-the-loop validation to address actuator dynamics and sensing latencies in real deployment scenarios.

\section*{ACKNOWLEDGMENT}
The authors thank Prof. Katsuo Kurabayashi and Prof. Rui Li for their guidance and our colleagues for their support. This project was supported by a research award from Nokia Bell Labs for studying embodied safety features for autonomous robotic systems. The authors acknowledge the institutional resources that enabled this work.

\bibliographystyle{IEEEtran}
\bibliography{references}

\begin{thebibliography}{10}
\providecommand{\url}[1]{#1}
\csname url@samestyle\endcsname
\providecommand{\newblock}{\relax}
\providecommand{\bibinfo}[2]{#2}
\providecommand{\BIBentrySTDinterwordspacing}{\spaceskip=0pt\relax}
\providecommand{\BIBentryALTinterwordstretchfactor}{4}
\providecommand{\BIBentryALTinterwordspacing}{\spaceskip=\fontdimen2\font plus
\BIBentryALTinterwordstretchfactor\fontdimen3\font minus \fontdimen4\font\relax}
\providecommand{\BIBforeignlanguage}[2]{{%
\expandafter\ifx\csname l@#1\endcsname\relax
\typeout{** WARNING: IEEEtran.bst: No hyphenation pattern has been}%
\typeout{** loaded for the language `#1'. Using the pattern for}%
\typeout{** the default language instead.}%
\else
\language=\csname l@#1\endcsname
\fi
#2}}
\providecommand{\BIBdecl}{\relax}
\BIBdecl

\bibitem{rudenko2020human}
A.~Rudenko, L.~Palmieri, M.~Herman, K.~M. Kitani, D.~M. Gavrila, and K.~O. Arras, ``Human motion trajectory prediction: A survey,'' \emph{International Journal of Robotics Research}, vol.~39, no.~8, pp. 895--935, 2020.

\bibitem{gui2018teaching}
L.-Y. Gui, K.~Zhang, Y.-X. Wang, X.~Liang, J.~M.~F. Moura, and M.~Veloso, ``Teaching robots to predict human motion,'' in \emph{Proceedings of the IEEE/RSJ International Conference on Intelligent Robots and Systems (IROS)}.\hskip 1em plus 0.5em minus 0.4em\relax IEEE, 2018, pp. 562--567.

\bibitem{luo2018unsupervised}
R.~Luo, R.~Hayne, and D.~Berenson, ``Unsupervised early prediction of human reaching for human--robot collaboration in shared workspaces,'' \emph{Autonomous Robots}, vol.~42, pp. 631--648, 2018.

\bibitem{luo2019human}
R.-C. Luo and L.~Mai, ``Human intention inference and online human hand motion prediction for human--robot collaboration,'' in \emph{Proceedings of the IEEE/RSJ International Conference on Intelligent Robots and Systems (IROS)}.\hskip 1em plus 0.5em minus 0.4em\relax IEEE, 2019, pp. 5958--5964.

\bibitem{lin2022human}
C.-H. Lin, K.-J. Wang, A.~A. Tadesse, and B.~H. Woldegiorgis, ``Human--robot collaboration empowered by hidden semi-markov model for operator behaviour prediction in a smart assembly system,'' \emph{Journal of Manufacturing Systems}, vol.~62, pp. 317--333, 2022.

\bibitem{liu2019deep}
Z.~Liu, Q.~Liu, W.~Xu, Z.~Liu, Z.~Zhou, and J.~Chen, ``Deep learning-based human motion prediction considering context awareness for human--robot collaboration in manufacturing,'' \emph{Procedia CIRP}, vol.~83, pp. 272--278, 2019.

\bibitem{butepage2018anticipating}
J.~B{\"u}t{\'e}page, H.~Kjellstr{\"o}m, and D.~Kragi{\'c}, ``Anticipating many futures: Online human motion prediction and generation for human--robot interaction,'' in \emph{Proceedings of the IEEE International Conference on Robotics and Automation (ICRA)}.\hskip 1em plus 0.5em minus 0.4em\relax IEEE, 2018, pp. 4563--4570.

\bibitem{kedia2024interct}
K.~Kedia, A.~Bhardwaj, P.~Dan, and S.~Choudhury, ``Interct: Transformer models for human intent prediction conditioned on robot actions,'' in \emph{Proceedings of the IEEE International Conference on Robotics and Automation (ICRA)}.\hskip 1em plus 0.5em minus 0.4em\relax IEEE, 2024.

\bibitem{zou2024simplified}
J.~Zou, B.~Chen, and C.~Lin, ``Simplified neural architecture for efficient human motion prediction in human--robot interaction,'' \emph{Neurocomputing}, 2024.

\bibitem{zhang2025prediction}
Y.~Zhang, G.~Peng, W.~Wang, Y.~Chen, Y.~Jia, and S.~Liu, ``Prediction-based human--robot collaboration in assembly tasks using a learning from demonstration model,'' \emph{Sensors}, 2025.

\bibitem{eltouny2023detgn}
M.~A. Eltouny, W.~Liu, S.~Tian, M.~Zheng, and X.~Liang, ``De-tgn: Uncertainty-aware human motion forecasting using deep ensembles,'' 2023, arXiv preprint.

\bibitem{halim2025hybrid}
M.~Y. Halim, L.~Canciani, and Q.~Zhu, ``Hybrid physics-infused deep learning for enhanced real-time human motion prediction in collaborative robotics,'' \emph{International Journal of Advanced Robotic Systems}, 2025.

\bibitem{pervez2008safe}
A.~Pervez and J.~Ryu, ``Safe physical human--robot interaction---past, present and future,'' \emph{Journal of Mechanical Science and Technology}, vol.~22, pp. 469--483, 2008.

\bibitem{mohammed2016active}
A.~Mohammed, B.~Schmidt, and L.~Wang, ``Active collision avoidance for human--robot collaboration driven by vision sensors,'' \emph{International Journal of Computer Integrated Manufacturing}, vol.~30, no.~9, pp. 970--980, 2016.

\bibitem{wang2025improved}
W.~Wang, L.~Zhao, and K.~Li, ``Improved artificial potential field method for redundant manipulator trajectory planning,'' \emph{Robotics and Autonomous Systems}, vol. 168, p. 104592, 2025.

\bibitem{guan2025improved}
T.~Guan, Y.~Han, M.~Kong, S.~Wang, D.~Feng, and W.~Yang, ``An improved artificial potential field with rrt* algorithm for autonomous vehicle path planning,'' \emph{Scientific Reports}, vol.~15, p. 16982, 2025.

\bibitem{chancharoen2022digital}
R.~Chancharoen, K.~Chaiprabha, L.~Wuttisittikulkij, W.~Asdornwised, M.~Saadi, and G.~Phanomchoeng, ``Digital twin for a collaborative painting robot,'' \emph{Sensors}, vol.~23, no.~1, p.~17, 2022.

\bibitem{hoebert2024framework}
T.~Hoebert, S.~Seibel, M.~Amersdorfer, M.~Vincze, W.~Lepuschitz, and M.~Merdan, ``A framework for enhanced human--robot collaboration during disassembly using digital twin and virtual reality,'' \emph{Robotics}, 2024.

\bibitem{inamura2025development}
T.~Inamura, H.~Yamada, K.~Morinaga, N.~Yamanobe, R.~Hanai, and Y.~Domae, ``Development and evaluation of a human--robot collaborative training system for retail stores using virtual reality and digital twin technologies,'' \emph{Journal of Robotics and Mechatronics}, vol.~37, no.~2, pp. 478--487, 2025.

\bibitem{malik2021digital}
T.~Malik and H.~S.~W. Seibel, ``Digital twins for collaborative robots: A case study in human--robot assembly,'' \emph{Journal of Robotics and Mechatronics}, 2021.

\bibitem{ramasubramanian2022digital}
A.~K. Ramasubramanian, R.~Mathew, M.~Kelly, V.~Hargaden, and N.~Papakostas, ``Digital twin for human--robot collaboration in manufacturing: Review and outlook,'' \emph{Applied Sciences}, vol.~12, no.~10, p. 4811, 2022.

\bibitem{baratta2024digital}
A.~Baratta, A.~Cimino, F.~Longo, and L.~Nicoletti, ``Digital twin for human--robot collaboration enhancement in manufacturing systems: Literature review and direction for future developments,'' \emph{Computers \& Industrial Engineering}, vol. 179, p. 109297, 2024.

\bibitem{elbasheer2022shaping}
M.~Elbasheer, F.~Longo, G.~Mirabelli, L.~Nicoletti, A.~Padovano, and V.~Solina, ``Shaping the role of digital twins for human--robot dyad: Connotations, scenarios, and future perspectives,'' \emph{Computational Intelligence in Manufacturing}, 2022.

\bibitem{langas2025exploring}
E.~F. Lang{\aa}s, ``Exploring the synergy of human--robot teaming, digital twins, and ai,'' \emph{Journal of Intelligent Manufacturing}, 2025.

\bibitem{langas2025synergy}
E.~F. Lang{\aa}s, M.~H. Zafar, and F.~Sanfilippo, ``Exploring the synergy of human--robot teaming, digital twins, and machine learning in industry 5.0: A step towards sustainable manufacturing,'' \emph{Journal of Intelligent Manufacturing}, 2025.

\bibitem{alham2023developing}
R.~Alham and M.~Hammadi, ``Developing a digital twin framework for monitoring the trajectory of the ur10 robot using an extended kalman filter,'' 2023, preprint.

\bibitem{chinnasamy2023digital}
S.~Chinnasamy, H.~Sura, A.~Saleem, A.~Kathirvel, and P.~Rangan, ``Digital twin of robot manipulator using ros,'' in \emph{AIP Conference Proceedings}, vol. 040004.\hskip 1em plus 0.5em minus 0.4em\relax AIP, 2023.

\bibitem{wang2025deep}
S.~Wang, H.~Zhang, J.~Liu, and C.~Wu, ``Deep learning-driven digital twin framework for human--robot collaboration,'' \emph{IEEE Transactions on Industrial Informatics}, vol.~21, no.~2, pp. 1234--1245, 2025.

\bibitem{kuts2017collaborative}
V.~Kuts, M.~Sarkans, T.~Otto, and T.~Tahemaa, ``Collaborative work between human and industrial robot in manufacturing by advanced safety monitoring system,'' \emph{International Journal of Automation and Smart Technology}, vol.~7, no.~4, pp. 179--192, 2017.

\bibitem{he2024development}
J.~He, Z.~Yan, Y.~Wang, Y.~Jiang, D.~Zeng, and Z.~Zhang, ``Development of robotic arm digital twins via edge-to-end architecture,'' in \emph{Proceedings of the IEEE}.\hskip 1em plus 0.5em minus 0.4em\relax IEEE, 2024.

\bibitem{renz2024moving}
H.~Renz, M.~Kr{\"a}mer, and T.~Bertram, ``Moving horizon planning for human--robot interaction,'' in \emph{Proceedings of the ACM/IEEE International Conference on Human-Robot Interaction (HRI)}.\hskip 1em plus 0.5em minus 0.4em\relax ACM/IEEE, 2024, pp. 112--120.

\bibitem{maruyama2021digital}
T.~Maruyama, T.~Ueshiba, M.~Tada, H.~Toda, Y.~Endo, Y.~Domae, Y.~Nakabo, T.~Mori, and K.~Suita, ``Digital twin-driven human--robot collaboration using a digital human,'' \emph{Sensors}, vol.~21, no.~24, p. 8266, 2021.

\bibitem{collins2021review}
A.~Collins, B.~Patel, C.~Huang, and D.~Lewis, ``A review of physics simulators for robotic applications,'' \emph{Journal of Robotics Simulation}, vol.~15, no.~2, pp. 101--115, 2021.

\bibitem{yoon2023comparative}
Y.~Yoon, H.~Kim, and J.~Park, ``Comparative study of physics engines for robot simulation with mechanical interaction,'' \emph{International Journal of Advanced Robotics}, vol.~30, no.~5, pp. 501--512, 2023.

\bibitem{cha2023human}
Y.~Cha, S.~Rhim, and J.~Kim, ``Human-aware trajectory planning algorithm using gru-based prediction and a-rrt*,'' in \emph{Proceedings of the IEEE International Conference on Robotics and Automation (ICRA)}.\hskip 1em plus 0.5em minus 0.4em\relax IEEE, 2023, pp. 2341--2348.

\bibitem{basei2025proactive}
E.~Basei, L.~Freda, and A.~Bicchi, ``Proactive motion planning for human--robot collaboration,'' in \emph{Proceedings of the IEEE International Conference on Robotics and Automation (ICRA), Workshop on Safe Collaborative Robotics}.\hskip 1em plus 0.5em minus 0.4em\relax IEEE, 2025.

\bibitem{sampieri2022pose}
A.~Sampieri, G.~D’Amely, A.~Avogaro, F.~Cunico, G.~Skenderi, F.~Setti, M.~Cristani, and F.~Galasso, ``Pose forecasting in industrial human--robot collaboration,'' in \emph{Proceedings of the European Conference on Computer Vision (ECCV)}.\hskip 1em plus 0.5em minus 0.4em\relax Springer, 2022.

\bibitem{guo2025dbvsb}
S.~Guo, Y.~Zhang, and H.~Wang, ``Dbvsb-p-rrt*: A bi-directional p-rrt* algorithm with adaptive direction bias and variable step-size for mobile robot path planning,'' \emph{Computers and Mathematics with Applications}, 2025.

\bibitem{collins2021whitepaper}
J.~Collins, S.~Chand, A.~Vanderkop, and D.~Howard, ``A review of physics simulators for robotic applications,'' 2021, unpublished white paper.

\bibitem{Chen2023_APF_JointSpace}
\BIBentryALTinterwordspacing
Y.~Chen, L.~Chen, J.~Ding, and Y.~Liu, ``Research on real-time obstacle avoidance motion planning of industrial robotic arm based on artificial potential field method in joint space,'' \emph{Applied Sciences}, vol.~13, p. 6973, 2023. [Online]. Available: \url{https://doi.org/10.3390/app13126973}
\BIBentrySTDinterwordspacing

\end{thebibliography}

\end{document}